%% file: iwslt2017-submission.tex
\documentclass[a4paper]{article}
\usepackage{iwslt17,amssymb,amsmath,epsfig}
\usepackage{times}
\usepackage{latexsym}
\usepackage{xspace}
\usepackage{graphicx}
\usepackage{amsmath}
\usepackage{mathtools}
\usepackage{algorithm2e}

\def\reg{{\rm\ooalign{\hfil
     \raise.07ex\hbox{\scriptsize R}\hfil\crcr\mathhexbox20D}}}

\input{defs}

\title{Synthetic  Data  for Neural Machine Translation of Spoken-Dialects}
 \makeatletter
 \def\name#1{\gdef\@name{#1\\}}
 \makeatother
 \name{{\em Hany Hassan, Mostafa Elaraby, Ahmed Y. Tawfik}}
\address{Microsoft AI \& Research  \\
 \\
{\small \tt {hanyh,a-moelar,atawfik}@microsoft.com}
}
\begin{document}
\maketitle

\maketitle

\begin{abstract}

	In this paper,  we introduce a novel approach to generate synthetic data for  training  Neural Machine Translation systems. The proposed approach supports language variants and dialects with very limited parallel training data. This is achieved using a seed data to project words  from a closely-related resource-rich language to an under-resourced language variant  via word embedding representations. The proposed approach is based on localized  embedding projection of distributed representations which utilizes monolingual embeddings and approximate nearest neighbors queries to transform parallel data across language variants.  
	
	Our approach is language independent and can be used to generate data for any variant of the source language such as slang or  spoken dialect  or even for a different language that is  related to the source language. We report experimental results on Levantine  to English translation using Neural Machine Translation. We show that the synthetic data can provide significant improvements over  a very large scale system by more than 2.8 Bleu points and it can be used to provide a reliable translation system for a spoken dialect which  does not have sufficient parallel data.

\end{abstract}

\section{Introduction}

Neural Machine Translation (NMT) \cite{BahdanauCB14} has achieved state-of-the-art translation quality  in various research  evaluations campaigns \cite{SennrichHB16a} and online large scale production systems \cite{WuSCLNMKCGMKSJL16} and \cite{jacob}. With such large systems,  NMT  showed  that it can scale up to huge amounts of parallel data. However, such large parallel data is not widely available for all domains and language styles. Usually parallel training data is widely available in written formal languages such as UN and Europarl data. 

Real-time speech translation systems support spontaneous, open-domain conversations between speakers of different languages. Speech Translation Systems are becoming a practical tool that can help in eliminating language barriers for spoken languages. 
Those machine translation systems are usually trained using NMT with large amount of parallel data adapted from written data to the spoken style \cite{umdiwslt}. This is a valid approach when the spoken and written languages are similar and mainly differ in style. For many languages, the written and spoken forms are quite different \textsection\ref{sec:spoken}. While the written form usually has an abundance of parallel data  available to train a reliable NMT system; the spoken form may not have any parallel data or even, in some cases, a standardized written form.

In this paper,  we propose a novel approach to generate synthetic data for NMT. The proposed approach  transforms a given parallel corpus between a  written language and a target language to a parallel corpus between the spoken dialect variant  and the target language. Our approach is language independent and can be used to generate data for any variant of the source language such as slang, spoken dialect or social media style or even for a different language that is closely-related to such source language.

  The synthetic data generation approach is based on two simple principles: first,  distributional word representation (word embeddings) can preserve similarity relations across languages \cite{MikolovSCCD13}. Secondly,  a localized  projection can be learned to transform between various representations \cite{ZhaoHA15}. We assume that we are trying to learn a translation system between $F^\prime$ and $E$, where $F^\prime$ is a variant of $F$, i.e. a spoken dialect. We start from parallel corpus between the two standard languages $F$ and $E$, then we transform it  into a three-way corpus between  $F$, $E$ and $F^\prime$.  The proposed approach assumes  the existence of  a seed bi-lingual lexicon or a small seed parallel data between $F^\prime$ and either $F$ or $E$.

The proposed approach is motivated by the assumption that both Language $F$ and its variant Language $F^\prime$ share  some vocabulary, have similar word orders and share similar bi-lingual characteristics with Language $E$. We start by constructing a continuous word representation (i.e. word2vec \cite{MikolovSCCD13})  for each one of the three languages. Using the seed bi-lexicon between either $E$ and $F^\prime$  or $F$ and $F^\prime$, we train a local projection to transform the words across the different representation spaces. 

We used the proposed approach to generate spoken Levantine-English data from Arabic-English data then we experimented with utilizing the generated data  in various settings to improve translation of the spoken dialect. The rest of this paper is organized as follows, Section \textsection\ref{sec:spoken} presents an overview of spoken dialects since it is the focus application of this work. Section \textsection\ref{sec:rel} discusses related work. Section \textsection\ref {sec:nmt} presents a brief overview of Neural machine translation. Section \textsection\ref{sec:datagen} discussed in detail the proposed approach for generating data. Section \textsection\ref{sec:exp} presents the experimental setup. Finally, we discuss the results and conclude in section \textsection\ref{sec:conc}.

\section{Spoken Language Variants}
\label{sec:spoken}

Some languages present an additional challenge to Spoken Language Translation (SLT) when the spoken variant differs significantly from the written one. Moreover, sometimes the spoken language used in the daily life is quite different than the standard form used in the education system as well as in formal communication such as news papers and  broadcast news. 
For example, Singapore English (Singlish)  is an English-based creole with a mix of English, Mandarin, Malay, and Tamil \cite{lim2010singapore}. Similarly, the  standard form of written Arabic is Modern Standard Arabic (MSA); however, it is not the spoken mother tongue by Arabic speakers.   The Arabic spoken dialects vary by geo-graphical region with at least five dialects: Egyptian, Levantine, Iraqi, gulf, and North African. While all dialects are stemmed from MSA, they are quite different  phonologically, lexically, morphologically and syntactically. For example, spoken colloquial Levantine Arabic conversations share between 61.7\% and 77.4\% of their vocabulary with a written news corpus from the same region \cite{Al-ShareefH11}. This results in spoken dialects that are quite different and not even well interpreted between Arabic speakers of different dialects. 

Most of the spoken language variants stem from a more formal written language such as Singlish from English and Levantine from MSA. While the spoken dialects do not usually have parallel data, they enjoy a wide adoption on social media which results in large monolingual corpora for such spoken variants. In this work, we are proposing a novel approach to overcome such limitation for spoken languages through generating parallel data leveraging the spoken dialects monolingual data and the written form parallel data.

In this paper we focus on Levantine-English translation as the pressing need for such translation systems due to the  refugee crisis that dictates  the need for  a reliable open-domain  translation  from Levantine to English. 


\section{Related Work}
\label{sec:rel}

There have been a number of proposed approaches to learn synthesized  translation units for statistical machine translation systems such as \cite{Klementiev}, \cite{SalujaHTQ14} and \cite{ZhaoHA15}. Such approaches focused on learning translation rules that would fit into a statistical phrase-based system. Those approaches do not fit into Neural Machine Translation (NMT) systems which require full context to learn to encode the sentences. 

A number of approaches  have been proposed utilizing monolingual target data into NMT training. Most notably, \cite{SennrichHB16} used  monolingual sentences by generating  pseudo parallel data through back-translating the monolingual data and using it in the reverse direction to improve NMT systems. Back-Translation  showed significant improvement especially in domain adaption setups.  The  back-translation approach  is not directly comparable to ours,  since ours does not require a pre-trained  system while back-translation does require one. However, we are using a seed parallel data as a source of our lexicon and it  would be fairly comparable to use such data in both settings as we report in our experiments. 

Dialectal Arabic translation has been a well-known problem; \cite{ZbibMDSMSMZC12} tried to solve this problem by crowd-sourcing translation for dialect data. They translated around (160K sentences) of Levantine and Egyptian data. The main limitation of this approach is that it is quite limited and not scalable. The vocabulary of the collected data is not sufficient to provide open-domain translation system. 
On the other hand, \cite{Durrani2014} and \cite{SajjadDGNAVSKH16} tried to solve the problem by applying rule-based transformation between Levantine or Egyptian to MSA. The main limitation of such approaches  is that they require extensive  linguistics knowledge to design the conversion rules which are not flexible to new vocabulary and styles that are constantly being introduced to the spoken languages.

\section{Neural Machine Translation}
\label{sec:nmt}
Neural Machine Translation is based on Sequence-to-Sequence encoder-decoder model as proposed in \cite{SutskeverVL14}  along with an attention mechanism to handle longer sentences \cite{BahdanauCB14} and \cite{LuongPM15}. 

In this work, we use an  in-house implementation \cite{jacob}  for attention-based encoder-decoder NMT which is similar to \cite{BahdanauCB14}. NMT is modeling the log conditional probability of the target sequence given the source as shown in eqn\ref{eq:prob}:

\begin{equation}
\log p(y|x) = \sum_{k=1}^n \log p(y_k|y_{<k},x)
\label{eq:prob}
\end{equation}

 NMT follows encoder-decoder architecture; the encoder is a bidirectional recurrent neural network (LSTM) that calculates the hidden encoder state at each word $h_1 h_2... h_m$. The decoder is another  recurrent neural network (LSTM) as well that calculates the hidden state at each decoded output state $s_1 s_2 .... s_n$. Then a softmax is applied to get a distribution over target words.

\begin{equation}
 y_k = softmax(g(y_{k-1},s_k, c_k))
\label{eq:y}
\end{equation}
where $c_k$ is calculated  by the attention mechanism which is a weighted sum of the encoder's hidden states that determines  the importance of each encoder  hidden state to the predicted output. The attention mechanism represents  the variable length input sequence as a weighted fixed-dimension context vector $c_k$
 
\begin{equation}
c_k = \sum_{i=1}^m \alpha_{ki} h_i
\label{eq:attn}
\end{equation}
where $\alpha_ki$ is calculated as a normalized weight of the association between the previous decoder state $s_{k-1}$ and the current encoder state $h_i$  which is calculated as a dot product as described in \cite{LuongPM15}.

During  training, all model's parameters are optimized jointly using stochastic gradient methods to maximize the conditional probability of all sentence pairs in the training data. At decoding time, one word is predicted at each step, a beam search is used to score the best translation path.

\section{Synthesized Data Generation}
\label{sec:datagen}

Our data generation approach is motivated by two observations: firstly, distributional  representations of words have been found to capture syntactic and semantic regularities in languages. In such continuous representation space, the relative positions between words are preserved across languages \cite{MikolovSCCD13}. Secondly, the representation spaces have localized sub-clusters of neighboring data points that form  smooth manifolds \cite{roweis2000nonlinear} which can be leveraged  to learn a localized transformation between the sub-clusters in different spaces across languages \cite{ZhaoHA15}. Since the sub-clusters are formed by similar words, a mapping can be learned between sub-clusters across representations. We exploit those characteristics to design our synthetic data generation approach.

The proposed  approach assumes the availability of three resources: (1) parallel data between Language $F$ and Language $E$,  (2) a seed lexicon or seed parallel data between either $E$ and $F^\prime$ or $F$ and $F^\prime$. (3) Monolingual corpora for $E$, $F$ and $F^\prime$ to train word vectors. The resulting synthesized data is a three-way data ($F$-$F^\prime$-$E$).  In this paper, we use a seed parallel data to acquire the lexicon between $E$ and $F^\prime$ through word alignment. However, a pre-existing lexicon can be used exactly the same way.

\begin{figure*}[tp]
	\centering
		\includegraphics[width=1.00\textwidth]{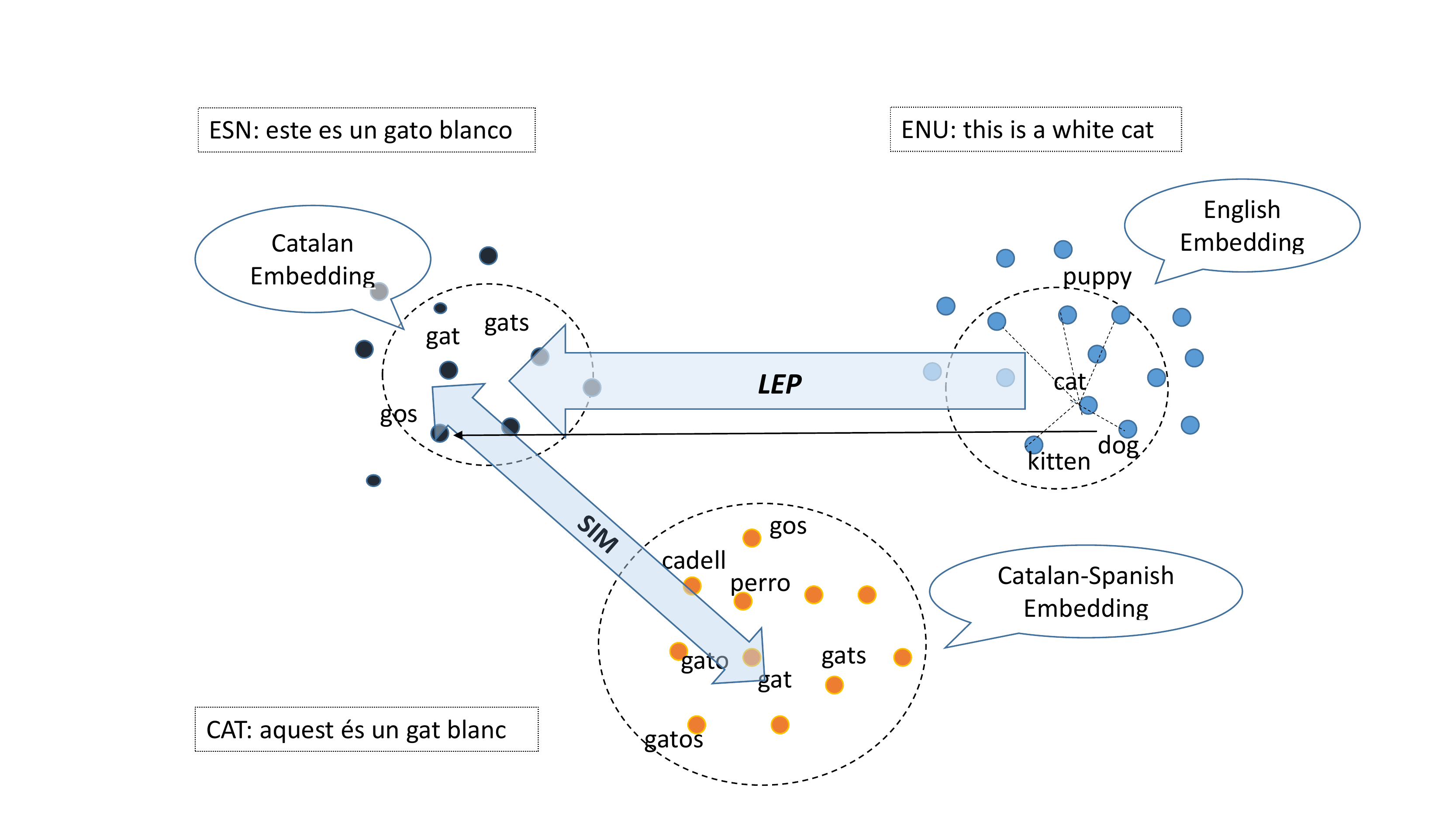}
	\caption{Synthetic Data Generation Using LEP}
	\label{fig:data-gen}
\end{figure*}

Figure \ref{fig:data-gen} illustrates the data generation process. For illustration purposes, let's assume that $E$ is English, $F$ is Spanish (ESN)  and we would like to generate $F^\prime$ which is Catalan (CAT) to English parallel data \footnote{The languages in the example are for illustration purposes only} . Furthermore, we assume that we have a seed lexicon between Catalan $F^\prime$  and English $E$ which we call  {\it BiLexicon}.

We build three distributional representations (i.e. word2vec)  using  monolingual corpora: the first is a  target representation for $E$, English in our example . The second is mixed source representation $F$-$F^\prime$ (Spanish-Catalan in our example). And the third is a Catalan ($F^\prime$) only embedding.

 The data generation  proceeds as follows:

\begin{itemize}
	\item For each English word $e$ in a sentence from $F$-$E$ parallel data, we query its k-nearest neighbors (\knn)
	\item \knn query on the $E$ embedding results  in a sub-cluster of $k$ English words around $e$. 
	\item If the $k$ queried neighbors do not contain at least $m$ words in {\it BiLexicon}, we repeat the query with $2k$.
	\item If no $m$ neighbors words can be retrieved,   the process  terminates for this word and move to the next word.
	\item We use $m$ to query {\it BiLexicon} for equivalent words in the  $F^\prime$ space.
	\item As shown in Figure \ref{fig:data-gen}, we use the two localized sub-clusters in $E$ (English) and $F^\prime$ (Catalan) spaces to learn a localized projection between the two spaces. This is done using Local Embedding Projection (LEP) \textsection\ref{sec:lep}.
	\item The locally trained LEP is used to project the current $E$ word $e$ to its equivalent vector in the $F^\prime$ space.
	\item We perform \knn query around the projected vector in the $F^\prime$ (Catalan) space  to get $n$ candidates words.
	\item We then rank the $n$ candidates words according to their similarity with the $F$ (Spanish) words $f$ aligned to the current English word $e$ based on word alignment of the $F$-$E$ parallel data.
	\item The similarity  is calculating cosine Similarity (SIM) in the Spanish-Catalan space between the candidate Catalan words and the Spanish word($f$). 
	\item The top ranked Catalan word $f^\prime$ is selected and substituted in palace of $f$
	\item Alternatively, we can obtain the alignment information between $E$ (English) and $F$ (Spanish) words either by conventional word alignment techniques or by using Bi-Lingual embeddings as described  in Section \textsection\ref{sec:reps}.
	
\end{itemize}

%


%
It is worth noting that  for one-to-many mappings, we construct a composed vector for the multiple words by performing addition  of their corresponding vectors. There are a few other approaches to compose multi-words vectors. However, it has been shown  empirically that simple additive method achieves good performance \cite{mitchell2010composition}.

Later on, we discuss  the  main components we utilize in the generation process: Word Representation \textsection\ref{sec:reps}, efficient Nearest Neighbors Search \textsection\ref{sec:knn}  and Local Embedding Projection \textsection\ref{sec:lep}.

\subsection {Word Representation}
\label{sec:reps}
Continuous representations of words have been found to capture syntactic and semantic regularities in languages \cite{MikolovSCCD13}.  The induced representations  tend to cluster similar words together. We  directly use continuous representations learned from monolingual corpora such as Continuous Bag-Of-Words (CBOW) representation. In such continuous
representation spaces, the relative positions between words are preserved across languages. 
As shown in Figure \ref{fig:data-gen}, we learn three independent representations for spoken source, target and mixed sources. Those can be learned from monolingual corpora using off-the-shelf tools such as word2vec \cite{MikolovSCCD13}.

We require a mapping between the words in the original parallel corpus which can be  obtained  by performing word alignment on the parallel sentences. Alternatively, this requirement can be relaxed by  using a bi-lingual embedding trained on any parallel corpus such as Bivec \cite{luong2015bilingual}. Instead of using word alignment to map the source word to target word(s), we initiate a query to bilingual representation to retrieve the most likely target  word mapped to a given source word. This can be handy in the case of using comparable corpus rather than parallel corpus. We evaluate the merit  of this approach in \textsection\ref{sec:bivec}

\subsection{Nearest Neighbors Search}
\label{sec:knn}

The algorithm discussed above, requires an extensive number of \knn queries per word, which are the most time-consuming part  of the procedure.  A brute force \knn query requires a linear search over the whole source or target vocabulary which is usually in the order of millions requiring   $O(n)$ search.  This dictates the need for a fast approximate \knn query technique. While such techniques are widely used in various machine learning areas especially in vision application, they are not well explored for text applications.

Approximated \knn query  usually involves two steps, an offline index construction step and an online query step. While the offline step does not affect the run-time, it can be memory consuming.
A good approximation sacrifices the query accuracy a little bit, but speeds up the query by orders of magnitude. Locality Sensitive Hashing (LSH) \cite{indyk1998approximate} is a popular technique, but its performance  decreases as the number of dimensions grows, therefore it is not a good match for high dimensional spaces like ours. In this paper, we use Multiple Random Projection Trees (MRPT) \cite{HyvonenPTJTWCR16} for approximated \knn queries.

MRPT \cite{HyvonenPTJTWCR16}  uses multiple random  projection  trees  to get a more  randomized
space-partitioning trees. The random projection trees  result in splitting  hyperplanes  that are  aligned  with   random  directions sampled  from  the  space hypersphere instead of the  coordinate  axes. Moreover, it utilizes voting search among the random projection trees to provide more randomization that leads to  fast  query  times  and  accurate  results. At run-time,  a query $\vecp$  is routed down in several trees, and then a   linear  search, similar to RBV, is  performed  in  the  union  of  the points of all the leaves the query point fells into, the result is the approximated $k$-nearest neighbors to $\vecp$.

\subsection{Localized Embedding Projection (LEP)}
\label{sec:lep}
The \knn queries result in two local clusters as shown in Figure~\ref{fig:data-gen}. Given a word in one of the sub-clusters we want to find similar word(s) in the corresponding target sub-cluster. We use  Localized Embedding Projection (LEP) to achieve this task.

LEP  is based on simple intuition: the two sub-clusters represent smooth manifolds where each data point in a sub-cluster can be mapped to a corresponding data point in the other sub-cluster using  local linear transformation.  LEP has been successfully used  in \cite{ZhaoHA15} to transform between various representations based on the {\it locally linear embedding} method which was originally  proposed in \cite{roweis2000nonlinear} for dimensionality reduction.

LEP utilizes a localized projection matrix for each word, this is unlike global linear projection, as proposed in \cite{MikolovSCCD13}, which uses a single projection matrix for the all words in the space. As shown in \cite{ZhaoHA15}, it can be brittle to small non-linearity in the representation vector space and therefore  it is not a good choice  for all possible words. Unlike global projection, local projection requires an additional \knn query to find the neighbors of each word.

In LEP,  a linear projection $W_\vecf$ is learned for each word $\vecf$  to map between its neighbors to the neighbors of the projected  points in the  projected/translation space.
$(\vecf_1, \vece_1), (\vecf_2, \vece_2), \dots, (\vecf_m, \vece_m)$, $\vecf_i\in N(\vecf)$.

Let's  denote $f$ and $e$ as source side and target side words
respectively, and $\vecf$ and $\vece$ as the corresponding words vectors. Following \cite{MikolovSCCD13},
we learn the linear projection $W$ 
from the translations of the $n$ most frequent labeled source side phrases:
$(\vecf_1, \vece_1), (\vecf_2, \vece_2), \dots, (\vecf_n, \vece_n)$. 
Denote $F=[\vecf_1^T, \vecf_2^T, \dots, \vecf_n^T]^T$, 
$E = [\vece_1^T, \vece_2^T, \dots, \vece_n^T]^T$.
$W$ is calculated by solving the following linear system:
\[FW = E,\]
whose solution is: 
\[W \approx(F^TF)^{-1}F^TE.\]

Once the linear transform $W$ is known,
for each word $\vecf$,
$\vecf W=\bar{\vece}$ is the location in the target side 
that should be close to the target words representing similar meaning.
A \knn query can fetch all the target word vectors near point $\bar{\vece}$.

\section{Experimental Setup}
\label{sec:exp}

We used the proposed approach to generate spoken Levantine-English data from Arabic-English data then we experimented with utilizing the generated data  in various settings to improve translation for the spoken dialect. 


\subsection{Datasets}
\label{sec:data}

The only publicly available Dialectal Arabic to English parallel corpus is LDC2012T09\footnote{https://catalog.ldc.upenn.edu/LDC2012T09} \cite{ZbibMDSMSMZC12}. It consists of about 160K sentences of web data of mixed Levantine and Egyptian manually translated to English. We use this data set as  our baseline and as a source for the seed lexicon between English and Levantine.

Our main focus is to develop an open-domain conversational   translation system for Levantine-English.  In recent translation evaluations, OpenSubtitles data \cite{TIEDEMANN12} has been found to yield good translation quality for conversational domains compared to other data sources \cite{umdiwslt}. Therefore, we opt for using OpenSubtitles-2013\footnote{http://opus.lingfil.uu.se/} which consists of ~3M sentences as our Arabic(MSA)-to-English parallel corpus, to generate Levantine-English Parallel corpus.

We have created a three-way test set to evaluate this work (LEV-ENG-Test), where the source is transcription of spontaneous Levantine audio conversations translated  into both English and MSA Arabic. The test set is composed of 6K sentences and has been used to report all results in this paper.

\begin{table}
\begin{center}
\begin{tabular}{lccc}
Corpus & English & Arabic MSA & Levantine \\
\hline
\# of Tokens & 2B & 1.1B & 106M\\
\# of Word Vectors & 5.1M & 6.8M &1.5M 
\end{tabular}
\end{center}
\caption{Monolingual corpora used in experiments.\label{tab:mono}}
\end{table}

We used monolingual corpora to  train three distributional representations of English, Levantine and Mixed (MSA with Levantine. The  data mostly consist of  Gigaword corpora, UN data, Subtitles  and web crawled data.
The information of these  corpora  is listed in Table ~\ref{tab:mono}.

After that we use the off-the-shelf  {\sf Word2Vec} \cite{MikolovSCCD13}
 to generate the word embeddings for each language using
the Continuous Bag-Of-Words scheme, where the number of dimensions $d=250$, 
$window=5$, $mincount=5$.

\subsection {Data filtering}
\label{sec:filter}
Our proposed approach depends on the quality of the  parallel data, we have noticed that OpenSubtitles data has a lot of misaligned or badly translated sentences. Therefore, we have trained a decision tree classifier to identify whether the sentence pair is noisy or not. We reject the sentence pairs that are noisy. The decision tree classifier utilizes features from the meta-data of the aligned sentence pairs, namely:  number of source words, number of target words,   unaligned percentage, length-normalized alignment confidence score and percentage of one-to-one alignments. We used 150 sentences manually  annotated to train the classifier with  Gini impurity with minimum samples split of 2 and minimum samples leaf of 1.

On the word level, we have applied a named entity tagger to detect named entities on either source or target sides to avoid mangling  them. We also used a stop-word list to avoid mapping them.

\subsection{NMT model and Pre-Processing}

Our NMT system is described in \textsection\ref{sec:nmt}, we use a bidirectional encoder with 1024-units LSTM and 2 layers decoder with attention. We use  embedding size of 512 and dropout of $0.2$. 

For pre-processing, we use Byte Pair Encoding PBE \cite{SennrichHB16a} with 32000 merging operations separately on the  source and target. This results in 35K source  and 34K target vocabularies. We limit the length of the sentences to 50 words. The training is done using Stochastic Gradient Descent (SGD) with Adam\cite{KingmaB14}.  We use mini-batch size of 64 and train for 1M steps. The translation quality is measured with lower-cased BLEU.

Across all experiments we use those hyper parameters for the data generation process described in \textsection\ref{sec:datagen}: $k=200$, $n=3$ and $m=5$.

\subsection{Bivec vs Word Alignment}
\label{sec:bivec}

In the first set of experiments, we have  evaluated whether we should use word-alignment information or Bivec \textsection\ref{sec:reps}  to connect the source and target words in the given parallel data. As shown in Table \ref{tab:rep}, our Baseline is trained on LDC2012T09 (160K)  of mostly Levantine-English data. We then generate 50K sentences from Arabic-English Subtitles data  with bilingual embedding (Gen-Bivec) and without it (Gen-Align). When we are not using Bivec, we just use the word alignment information on the Arabic-English parallel corpus to get the mapping between the words. The result shows that using alignment information is better than using Bivec in this case. It worth noting that using  Bivec may be handy if the data is comparable  data. In the rest of this work we  used word alignment information since it yields better performance.

\begin{table}
{\small
\begin{center}
\begin{tabular}{c|c|c}
System & Data Size & B LEV-ENG-Test \\
\hline
Baseline & 160K & 16.15 \\
Gen-BiVec & 210K & 16.43  \\
Gen-Align & 210K & 16.98   \\ 
\end{tabular}
\end{center}
\caption{Translation performances using BLEU on LEV-ENG-Test for using Bivec vs. word alignment 
\label{tab:rep}}
}
\end{table}

\subsection{Data Generation Experiments}

In this set of experiments,  we added more generated data from the subtitles data applying the filtering described above \textsection\ref{sec:filter}. We end up with 1.1M sentences candidates for generation which we use for generating LEV-ENG data. In this setup, we also compared our approach with back-translation \cite{SennrichHB16} which is commonly used with NMT.  The  back-translation  is not directly comparable to ours,  since ours does not require a pre-trained  system while back-translation does require one. However, we are using a seed parallel data as a source of our lexicon and it  would be fairly comparable to use such data in both settings.

Furthermore, we investigated two different models to utilize the synthetic data. The first just used the LEV-ENG data while the second leveraged the 3-way characteristic of the generated corpus LEV-MSA-ENG.

We train the following systems:
\begin{itemize}
	\item Baseline: This is trained on LDC Levantine-English corpus of 160K. Which is also part of all other systems reported below.
	\item Baseline-PBMT: this is the same as above but trained as a phrase-based system, following standard practice.
	\item Baseline-MSA: This is trained on LDC data in addition to 1.1M sentence pairs of filtered subtitles data which is MSA-English.
	\item BT: We  trained an English-Levantine system similar to the Baseline though in the reverse direction; we used it to back-translate the 1.1M subtitles data from English into Levantine.
	\item Gen-Mono-1M: This is the system using the generated LEV-ENG data.
	\item LEV-MSA--MSA-ENG: This a pipeline system where we train a system to convert LEV to MSA using the 3-way generated data, followed by MSA to ENG translation.
\end{itemize}

\begin{table}
{\small
\begin{center}
\begin{tabular}{c|c}
System & BLEU on LEV-ENG-Test\\
\hline
Baseline & 16.15 \\
Baseline-PBMT & 16.42 \\
Baseline-MSA & 15.37  \\ 
BT &  16.59 \\
Gen-Mono &  17.33\\
LEV-MSA--MSA-ENG &  12.87\\

\end{tabular}
\end{center}
\caption{Translation performances in BLEU for NMT with Generated data
\label{tab:nmt1}}
}
\end{table}

Table \ref{tab:nmt1} shows that adding the MSA subtitles data (Baseline-MSA) hurt the performance, this is quite expected since the data is mainly MSA but it add a fair comparison in terms of the size of the training data. The phrase-based baseline is slightly better than NMT baseline as expected in such low resource case.

Back-translation helped a little bit ( ~0.3 BLEU), we think the system trained on LDC parallel data is quite small to provide good lexical coverage to generate variates of  the translated data that can help in back-translation.

Adding the synthetic data (Gen-Mono) is quite useful and improves the performance by more than 2 BLEU points. Compared to back-translation, the synthesized data utilized  monolingual representation which can lead to lexical varieties that help in having better translation examples.

Since the generated data is a 3-way corpus LEV-MSA-ENG, we can leverage this by training a system that translates from LEV to MSA. At run-time, we use a pipeline of two systems: LEV-MSA followed by MSA-ENG. We experimented with two variants of LEV-to-MSA system, subwords-based and character-based. We found out that the system is not producing reasonable results since it produces MSA words not related to the LEV words in input.  We think one  reason is that MSA and LEV shares a lot of their vocabulary together; in our monolingual data sets listed in Table ~\ref{tab:mono}, they share 58\% of their vocabulary. The system tends to replace MSA words (in LEV input) to other MSA words. The resulted outcome is very noisy MSA sentences that not closely related to the LEV input.

\subsection{Open-domain NMT System Experiments}
Our main objective in this work is to enable large scale NMT systems to support spoken dialects. Therefore, we experimented  with a very large scale Arabic-English open-domain system trying to adapt it to Levantine using the synthetic  data. The large scale system uses UN data,  subtitles data and various web crawled data with a total of ~42M parallel sentences. The system is an ensemble of two identical systems that only differ by initialization, each ensemble is trained for 10 epochs on the data. We tried two approaches to utilize the synthetic data: adding it to the training data as usual and adapting one of the two ensembles by continuing to train it on the synthetic data for 2 more epochs, similar to the approach proposed in \cite{FreitagA16}.

 For this set of experiments, we have added 2M synthetic  Levantine-English sentences.
We also report results on NIST-08 Arabic-English which is a 4-references test-set \footnote{https://catalog.ldc.upenn.edu/LDC2010T01}. Furthermore, we report results on the human converted LEV-MSA-ENG which is the same as LEV-ENG-Test testset but translated into MSA as well by human annotators.  Since LEV data is converted to MSA by annotators, translating the human-converted test set can  represent the oracle score that we can get using an MSA trained system on this test set. This would help us understand how good the system using the generated data compared to MSA systems.

\begin{table}
{\small
\begin{center}
\begin{tabular}{c|c|c|c}
System & LEV-ENG-Test  & LEV-MSA-ENG & NIST08 \\
\hline
Large-Sys & 25.03  & 28.20 & 53.45 \\
Large+GenData & 27.91 &  28.32 & 53.42 \\
Large+Adapted & 27.37 & 27.45 & 52.97  \\ 
\end{tabular}
\end{center}
\caption{Translation performances in BLEU for Large Scale NMT with Generated data
\label{tab:nmt-lg}}
}
\end{table}

\begin{figure*}[t]
	\centering
		\includegraphics[width=1.00\textwidth, trim=0 540 0 50, clip]{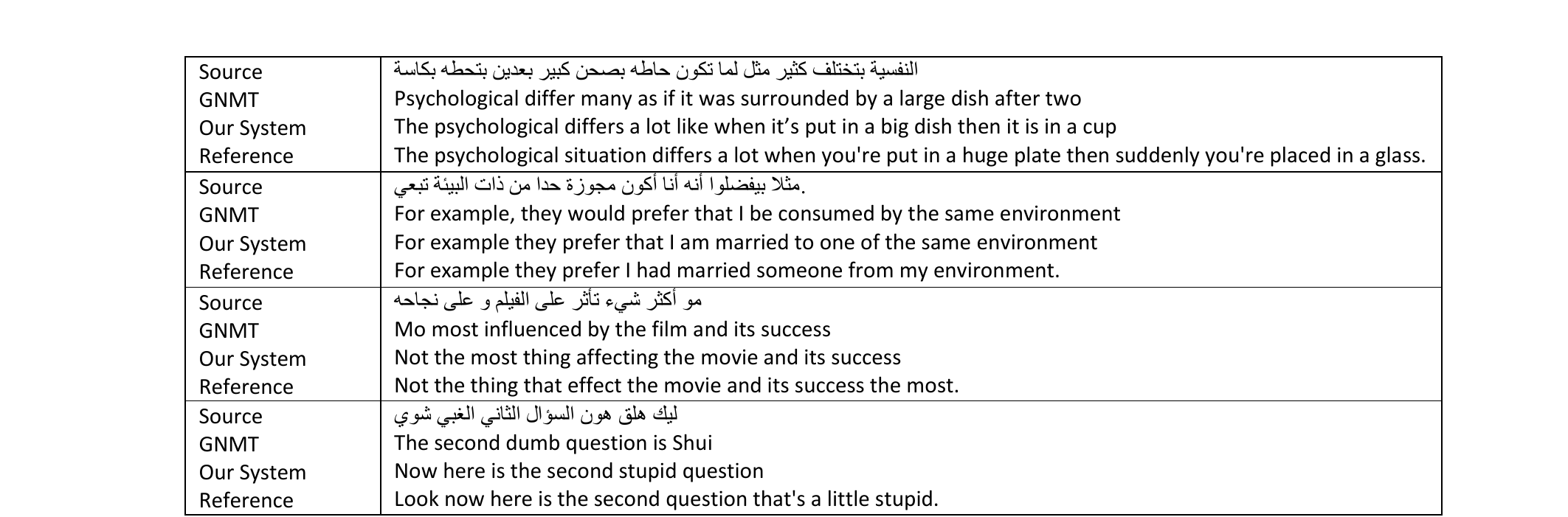}
	\label{fig:example-compare}
	\caption{Examples of system output and comparison}
\end{figure*}

As shown in Table \ref{tab:nmt-lg}, we get a very good improvement when adding the synthetic data as additional training data (Large+GenData) with 2.8 BLEU points. The performance of the system with the synthetic data is just 0.3 BLEU less than the oracle score on the human translated MSA (27.91 vs 28.20). Moreover, the addition of the synthetic data did not negatively affect the MSA NIST08 test sets as well; this simply enables us to have a single system to serve both written and spoken variants. This is a nice characteristic of NMT systems where  encoders can successfully handle varieties of source data as has been utilized in multi-lingual systems \cite{FiratCB16}.

   Adapting the system did help as well but not as good as re-training from scratch, however it may be a good option to avoid retraining the large system again.

Figure \ref{fig:example-compare} shows some cherry picked examples that show the improvement of the proposed  approach compared to GNMT \cite{WuSCLNMKCGMKSJL16} online neural system. It is quite clear that our system is doing much better compared to a large scale neural system.

\section{Discussion and Conclusion}
\label{sec:conc}
In this paper we presented a novel approach for generating  synthetic parallel  data for spoken dialects to overcome the limitations of the training data availability for such language variants. We show that we need to start from a corresponding parallel data and a seed lexicon or small parallel data. The results show that this approach is quite efficient and useful to improve general purpose NMT systems to the spoken variants. 

As for the future work, we would like to investigate the utilization of this approach for more languages as well as  different variants such as social media text translation. As a further step, we are  investigating the possibility of training the transformation process end-to-end within the neural machine translation system using a single neural network through learning the transformation from the sample seeds while making use of the monolingual corpus to learn the embeddings. 

\section{Acknowledgments}
We would like to thank Christian Federman and Will Lewis for fruitful discussions and helping in creating the test set.


\bibliographystyle{IEEEtran}

\end{document}

%% file: defs.tex
\newcommand{\knn}{$k$-NN\xspace}
\newcommand{\vecp}{\ensuremath{\mathbf{p}}\xspace}

\newcommand{\vecf}{\ensuremath{\mathbf{f}}\xspace}
\newcommand{\vece}{\ensuremath{\mathbf{e}}\xspace}
